# A Recommender System based on the Immune Network

Steve Cayzer[1] and Uwe Aickelin[2]
[1]Hewlett-Packard Laboratories, Filton Road, BS12 6QZ Bristol, UK, steve_cayzer@hp.com
[2] School of Computer Science, University of Nottingham, NG8 1BB UK, uxa@cs.nott.ac.uk

**Abstract-The immune system is a complex biological system with a highly distributed, adaptive and self-organising nature. This paper presents an artificial immune system (AIS) that exploits some of these characteristics and is applied to the task of film recommendation by collaborative filtering (CF). Natural evolution and in particular the immune system have not been designed for classical optimisation. However, for this problem, we are not interested in finding a single optimum. Rather we intend to identify a sub-set of good matches on which recommendations can be based. It is our hypothesis that an AIS built on two central aspects of the biological immune system will be an ideal candidate to achieve this: Antigen - antibody interaction for matching and antibody - antibody interaction for diversity. Computational results are presented in support of this conjecture and compared to those found by other CF techniques.**

## I. INTRODUCTION

Over the last few years, a novel computational intelligence technique, inspired by biology, has emerged: the artificial immune system (AIS). This section introduces the AIS and shows how it can be used for solving computational problems. In essence, the immune system is used here as inspiration to create an unsupervised machine-learning algorithm. The immune system metaphor will be explored, involving a brief overview of the basic immunological theories that are relevant to our work. We also introduce the basic concepts of collaborative filtering (CF).

*Overview of the Immune System*

A detailed overview of the immune system can be found in many textbooks [14]. Briefly, the purpose of the immune system is to protect the body against infection and includes a set of mechanisms collectively termed humoral immunity. This refers to a population of circulating white blood cells called B-lymphocytes, and the antibodies they create.

The features that are particularly relevant to our research are matching, diversity and distributed control. Matching refers to the binding between antibodies and antigens. Diversity refers to the fact that, in order to achieve optimal antigen space coverage, antibody diversity must be encouraged [11]. Distributed control means that there is no central controller, rather, the immune system is governed by local interactions between cells and antibodies.

The idiotypic network hypothesis [13] (disputed by some immunologists) builds on the recognition that antibodies can match other antibodies as well as antigens. Hence, an antibody may be matched by other antibodies, which in turn may be matched by yet other antibodies. This activation can continue to spread through the population and potentially has much explanatory power. The idiotypic network has been formalised by a number of theoretical immunologists [15].

There are many more features of the immune system, including adaptation, immunological memory and protection against auto-immune attack. Since these are not directly relevant to this work, they will not be reviewed here.

*Overview of Collaborative Filtering*

In this paper, we are using an AIS as a CF technique. CF is the term for a broad range of algorithms that use similarity measures to obtain recommendations. The best-known example is probably the "people who bought this also bought" feature of the internet company Amazon [2]. However, any problem domain where users are required to rate items is amenable to CF techniques. Commercial applications are usually called recommender systems [16]. A canonical example is movie recommendation.

In traditional CF, the items to be recommended are treated as 'black boxes'. That is, your recommendations are based purely on the votes of your neighbours, and not on the content of the item. The preferences of a user, usually a set of votes on an item, comprise a user profile, and these profiles are compared to build a neighbourhood. The key decisions to be made are:

Data encoding: Perhaps the most obvious representation for a user profile is a string of numbers, where the length is the number of items, and the position is the item identifier. Each number represents the 'vote' for an item. Votes are sometimes binary (e.g. did you visit this web page?) but can also be integers in a range (say [0,5]) or rational numbers.

Similarity Measure: The most common method to compare two users is a correlation-based measure like Pearson or Spearman, which gives two neighbours a matching score between -1 and 1. Vector based, e.g. cosine of the angle between vectors, and probabilistic methods are alternative approaches.

The canonical example is the k Nearest Neighbour algorithm, which uses a matching method to select k reviewers with high similarity measures. The votes from these reviewers, suitably weighted, are used to make predictions and recommendations.

Many improvements on this method are possible [10]. For example, the user profiles are usually extremely sparse because many items are not rated. This means that similarity measurements are both inefficient (the so-called 'curse of dimensionality') and difficult to calculate due to the small overlap. Default votes are sometimes used for items a user has not explicitly voted on, and these can increase the overlap size [4]. Dimensionality reduction methods, such as Single Value Decomposition, both improve efficiency and increase overlap [3]. Other pre-processing methods are often used, e.g. clustering [1]. Content-based information can be used to enhance the pure CF approach [10], [6]. Finally, the weighting of each neighbour can be adjusted by training, and there are many learning algorithms available for this [7]. All these improvements could in principle be applied to our AIS but in the interests of a clear and uncluttered comparison we have kept the CF algorithm as simple as possible.

The evaluation of a CF algorithm usually centres on its accuracy. There is a difference between prediction (given a movie, predict a given user's rating of that movie) and recommendation (given a user, suggest movies that are likely to attract a high rating). Prediction is easier to assess quantitatively but recommendation is a more natural fit to the movie domain. We present results evaluating both these behaviours.

*Using an AIS for Collaborative Filtering*

To us, the attraction of the immune system is this: if an adaptive pool of antibodies can produce 'intelligent' behaviour, can we harness the power of this computation to tackle the problem of preference matching and recommendation? Thus, in the first instance we intend to build a model where known user preferences are our pool of antibodies and the new preferences to be matched is the antigen in question.

Our conjecture is that if the concentrations of those antibodies that provide a better match are allowed to increase over time, we should end up with a subset of good matches. However, we are not interested in optimising, i.e. in finding the one best match. Instead, we require a set of antibodies that are a close match but which at the same time distinct from each other for successful recommendation. This is where we propose to harness the idiotypic effects of binding antibodies to similar antibodies to encourage diversity.

The next section presents more details of our problem and explains the AIS model we intend to use. We then describe the experimental set-up and present some initial results. Finally we review the results and discuss some possibilities for future work.

## 2. ALGORITHMS

*Application of the AIS to the EachMovie Tasks*

The EachMovie database [5] is a public database, which records explicit votes of users for movies. It holds 2,811,983 votes taken from 72,916 users on 1,628 films. The task is to use this data to make predictions and recommendations. In the former case, we provide an estimated vote for a previously unseen movie. In the latter case, we present a ranked list of movies that the user might like.

The basic approach of CF, is to use information from a neighbourhood to make useful predictions and recommendations. The central task we set ourselves is to identify a suitable neighbourhood. The SWAMI (Shared Wisdom through the Amalgamation of Many Interpretations) framework [9] is a publicly accessible software for CF experiments. Its central algorithm is as follows:

```
Select a set of test users randomly from the database
FOR each test user t
    Reserve a vote of this user, i.e. hide from predictor)
    From remaining votes create a new training user t'
    Select neighbourhood of k reviewers based on t'
    Use neighbourhood to predict vote
    Compare this with actual vote and collect statistics
NEXT t
```

The code shown in *italics* indicates a place where SWAMI allows an implementation-dependent choice of algorithm. We use an AIS to perform selection and prediction as below.

*Algorithm Choices*

We use the SWAMI data encoding:

$$User = \{\{id_1, score_1\}, \{id_2, score_2\} ... \{id_n, score_n\}\}$$

Where *id* corresponds to the unique identifier of the movie being rated and score to this user's score for that movie. This captures the essential features of the data available.

EachMovie vote data links a person with a movie and assigns a score (taken from the set {0, 0.2, 0.4, 0.6, 0.8, 1.0} where 0 is the worst). User demographic information (e.g. age and gender) is provided but this is not used in our encoding. Content information about movies (e.g. category) is similarly not used.

*Similarity Measure*

The Pearson measure is used to compare two users *u* and *v*:

$$r = \frac{\sum_{i=1}^{n} (u_i - \bar{u})(v_i - \bar{v})}{\sqrt{\sum_{i=1}^{n} (u_i - \bar{u})^2 \sum_{i=1}^{n} (v_i - \bar{v})^2}} \qquad (1)$$

Where *u* and *v* are users, *n* is the number of overlapping votes (i.e. movies for which both *u* and *v* have voted), $u_i$ is the vote of user *u* for movie *i* and $\bar{u}$ is the average vote of user *u* over all films (not just the overlapping votes). The measure is amended as follows

$$\begin{aligned} &if\ n = 0,\ \ r = NoOverlapDefault \\ &if \sum_{i=1}^{n} (u_i - \bar{u})^2 \sum_{i=1}^{n} (v_i - \bar{v})^2 = 0,\ \ r = ZeroVarianceDefault \qquad (2) \\ &if\ n < P,\ \ r = \frac{n}{P} r \qquad (where\ \ P = overlap\ penalty) \end{aligned}$$

The two default values are required because it is impossible to calculate a Pearson measure in such cases. Both were set to 0. Some experimentation showed that an overlap penalty *P* was beneficial (this lowers the absolute correlation for users with only a small overlap) but that the exact value was not critical. We choose a value of 100 because this is the maximum overlap expected.

*Neighbourhood Selection*

For a Simple Pearson predictor, neighbourhood selection means simply choosing the best $k$ (absolute) correlation scores, where $k$ is the neighbourhood size. Not every potential neighbour will have rated the film to be predicted. Reviewers who did not vote on the film are not added to the neighbourhood.

For the AIS predictor, a more involved procedure is required:

```
Initialise AIS
Encode user for whom to make predictions as antigen Ag
WHILE (AIS not stabilised) & (Reviewers available) DO
   Add next user as an antibody Ab
   Calculate matching scores between Ab and Ag
   Calculate matching scores between Ab and other antibodies
   WHILE (AIS at full size) & (AIS not stable) DO
         Iterate AIS
   OD
OD
```

Our AIS behaves as follows: At each step (iteration) an antibody's concentration is increased by an amount dependent on its matching to the antigen and decreased by an amount which depends on its matching to other antibodies. In absence of either, an antibody's concentration will slowly decrease over time. Antibodies with a sufficiently low concentration are removed from the system, whereas antibodies with a high concentration may saturate. An AIS iteration is governed by the following equation:

$$\frac{dx_i}{dt} = \begin{pmatrix} antigen \\ stimulation \end{pmatrix} - \begin{pmatrix} antibody \\ su\ ppression \end{pmatrix} - \begin{pmatrix} death \\ rate \end{pmatrix}$$
$$= k_1 m_i x_i y - \frac{k_2}{N}\sum_{j-1}^{N} m_{ij} x_i x_j - k_3 x_i \qquad (3)$$

$$m_{ij} = |r|$$
$$k_1 = Stimulation,\ k_2 = Suppression,\ k_3 = Death\ Rate$$
$$N = Number\ antibodies$$
$$x_i\,(or\ y) = concentration\ of\ antibody\ (or\ antigen)$$

This is a slightly modified version of Farmer et al's equation [8]. In particular, the first term is simplified as we only have one antigen, and we normalise the suppression term to allow a 'like for like' comparison between the different rate constants. $k_1$ and $k_2$ were varied as described in the next section. $k_3$ was fixed at 0.1, while the concentration range was set at 0–100 (initially 10). We fixed $N$ at 100. The matching function is the absolute value of the Pearson correlation measure. This allows us to have both positively and negatively correlated users in our neighbourhood, which increases the pool of neighbours available to us.

The AIS is considered stable after iterating for ten iterations without changing in size. Stabilisation thus means that a sufficient number of 'good' neighbours have been identified and therefore a prediction can be made. 'Poor' neighbours would be expected to drop out of the AIS after a few iterations.

Once the AIS has stabilised using the above algorithm, we use the antibody concentration to weigh the neighbours. However, early experiments showed that the most recently added antibodies were at a disadvantage compared to earlier antibodies. This is because they have had no time to mature (i.e. increase in concentration). Likewise, the earliest antibodies had saturated. To overcome this, we reset the concentrations and allow a limited run of the AIS to differentiate the concentrations:

```
Reset AIS (set all antibodies to initial concentrations)
WHILE (No antibody at maximum concentration) DO
   Iterate AIS
OD
```

*Prediction*

We predict a rating $p_i$ by using a weighted average over $N$, the neighbourhood of $u$, which was taken as the entire AIS.

$$p_i = \bar{u} + \frac{\sum_{v \in N} w_{uv}(v_i - \bar{v})}{\sum_{v \in N} w_{uv}} \qquad (4)$$

$$w_{uv} = r_{uv} x_v \qquad (NB\ relative\ not\ absolute)$$

Where $w_{uv}$ is the weight between users $u$ and $v$, $r_{uv}$ is the correlation score between $u$ and $v$, and $x_v$ is the concentration of the antibody corresponding to user $v$.

*Evaluation*

<u>Prediction Accuracy</u>: We take the mean absolute error, where $n_p$ is the number of predictions:

$$MAE = \frac{\sum |actual - predicted|}{n_p} \qquad (5)$$

<u>Mean number of recommendations</u>: This is the total number of unique films rated by the neighbours.

<u>Mean overlap size</u>: This is the number of recommendations that the user has also seen.

<u>Mean accuracy of recommendations</u>: Each overlapped film has an actual vote (from the antigen) and a predicted vote (from the neighbours). The overlapped films were ranked on both actual and predicted vote, breaking ties by movie ID. The two ranked lists were compared using Kendall's Tau τ. This measure reflects the level of concordance in the lists by counting the number of discordant pairs. To do this we order the films by vote and apply the following formulae:

$$\tau = 1 - \frac{4N_D}{n(n-1)}$$
$$N_D = \sum_{i=1}^{n} \sum_{j=i+1}^{n} D(r_i, r_j) \qquad (6)$$
$$D(r_i, r_j) = \begin{cases} 1 \, if \; r_i > r_j \\ 0 \, otherwise \end{cases}$$

Where $n$ is the overlap size and $r_i$ is the rank of film $i$ as recommended by the neighbourhood. Note that i here refers to the antigen rank of the film, not the film ID. $N_D$ is the number of discordant pairs, or, equivalently, the expected cost of a bubble sort to reconcile the two lists. $D$ is set to one if the rankings are discordant.

Mean number of reviewers. This is the number of reviewers looked at before the AIS stabilised.

Mean number of neighbours: This is the final number of neighbours in the stabilised AIS.

## 3 EXPERIMENTS

Experiments were carried out on a Pentium 700 with 256MB RAM, running Windows 2000. The AIS was coded in Java™ JDK1.3. Each run involved looking at up to 15,000 reviewers (20% of the EachMovie data set, randomly chosen) to provide predictions and recommendations for 100 users. Averaged statistics are then taken for each run. Runtimes ranged from 5 to 60 minutes, largely dependent on the number of reviewers.

*Experiments on Simple AIS*

Initial experiments concentrated on a simple AIS, with no idiotypic effects. The goal was to find a good stimulation rate, but also to ensure that the 'baseline' system operates similarly to a Simple Pearson predictor (SP). Therefore, we set the suppression rate to zero, and varied only the stimulation rate, i.e. the weighting given to antigen binding. Other parameters had been fixed by preliminary experiments.

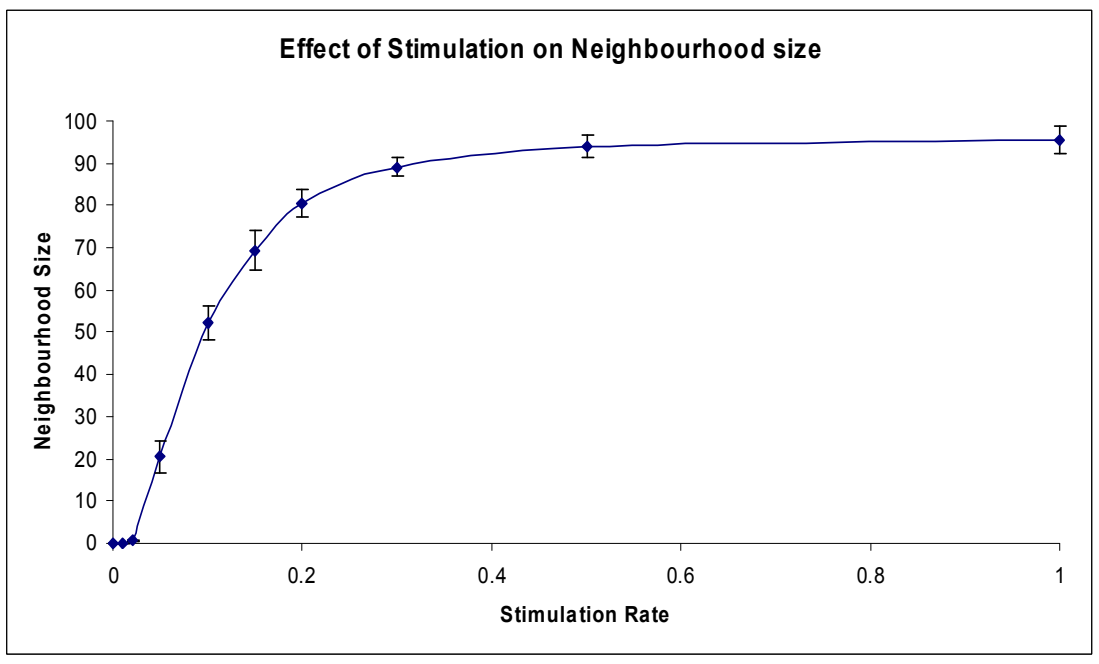

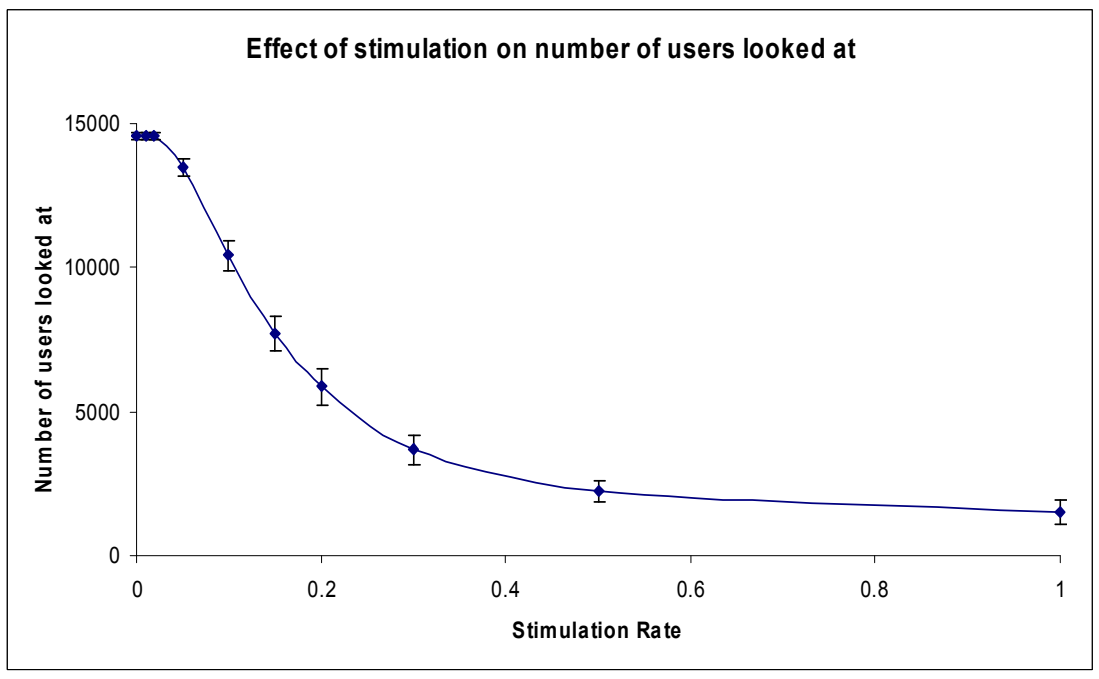

Figure 1: Effect of stimulation rate on neighbourhood and reviewers.

The graphs show averaged results over five runs at each stimulation rate. The bars show standard deviations. In order to have a fair comparison, the Simple Pearson parameters (neighbourhood and number of reviewers looked at) match the AIS values for each rate. In figure 2, we show the prediction error, number of recommendations, number of overlaps and recommendation accuracy for each algorithm. Note that low prediction error values are better, whereas for the other measures we are looking for high values.

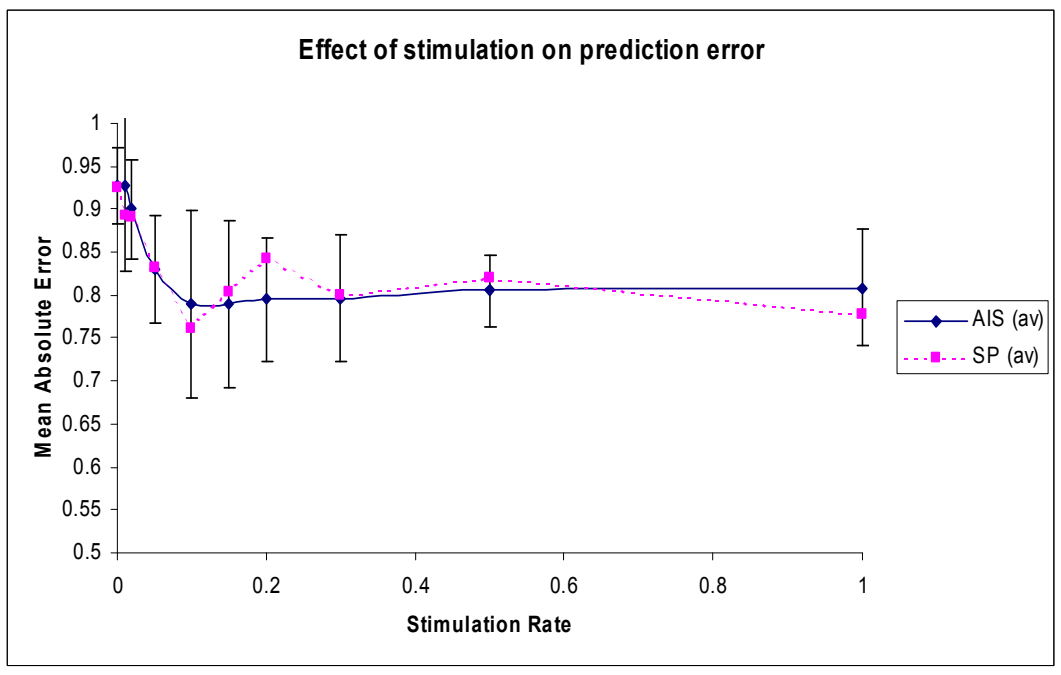

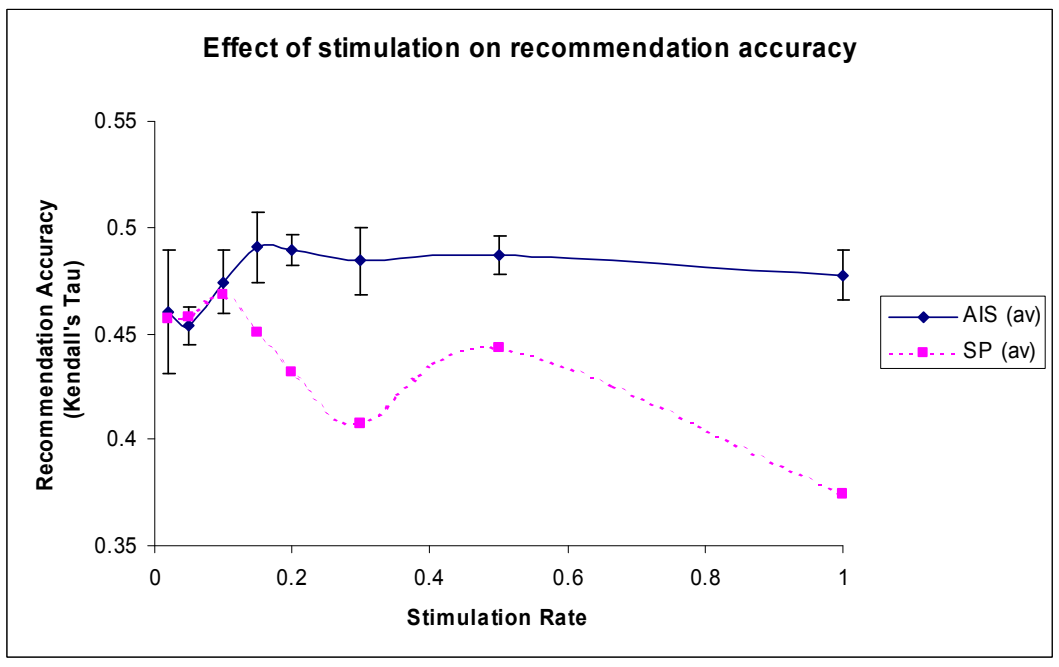

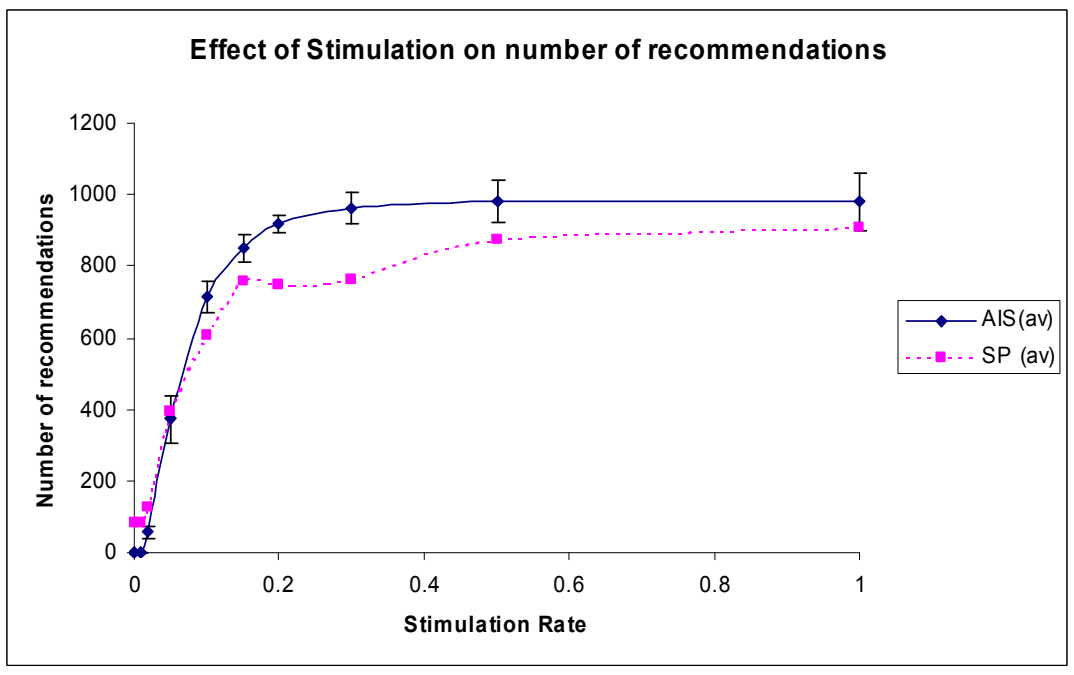

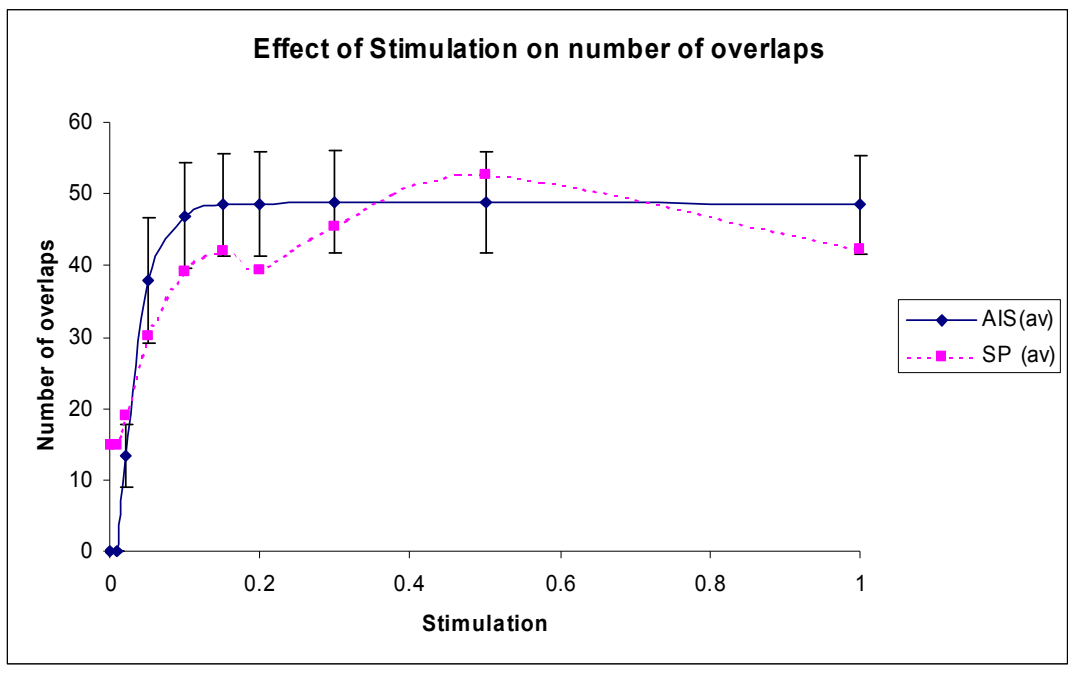

Figure 2: Effect of stimulation rate on prediction and recommendation.

It can be seen that the simple AIS gives broadly similar prediction performance to the Simple Pearson. The MAE measurements from different runs are not normally distributed, so a non-parametric statistic is appropriate. We performed a Wilcoxon analysis, which showed that the difference between prediction errors of SP and AIS is zero with 95% confidence. In addition, the choice of an appropriate stimulation rate did make a significant difference (a rate of 0.2 compared with 0.02 at the 95% level).

For recommendation, the AIS performs better than the SP at stimulation rates above 0.1. Again, we performed a positive 95% Wilcoxon analysis to assess significance. We excluded cases where a recommendation score was unavailable (due to an insufficient number of overlaps). The number of recommendations and overlaps show similar trends though the AIS gives a more constant value. Again, some stimulation was beneficial.

In later experiments, the stimulation rate was fixed at one of the better values (0.2, 0.3 or 0.5), in order to give us a good base to work on. These values give us generally good performance, while keeping a good neighbourhood size and still evaluating a reasonable number of reviewers.

*Experiments on the Idiotypic AIS*

Having fixed all the simple parameters, we tested the effect of suppression for stimulation rates of 0.2, 0.3 and 0.5. Not surprisingly we found that suppression changed the number of reviewers looked at and the number of neighbours:

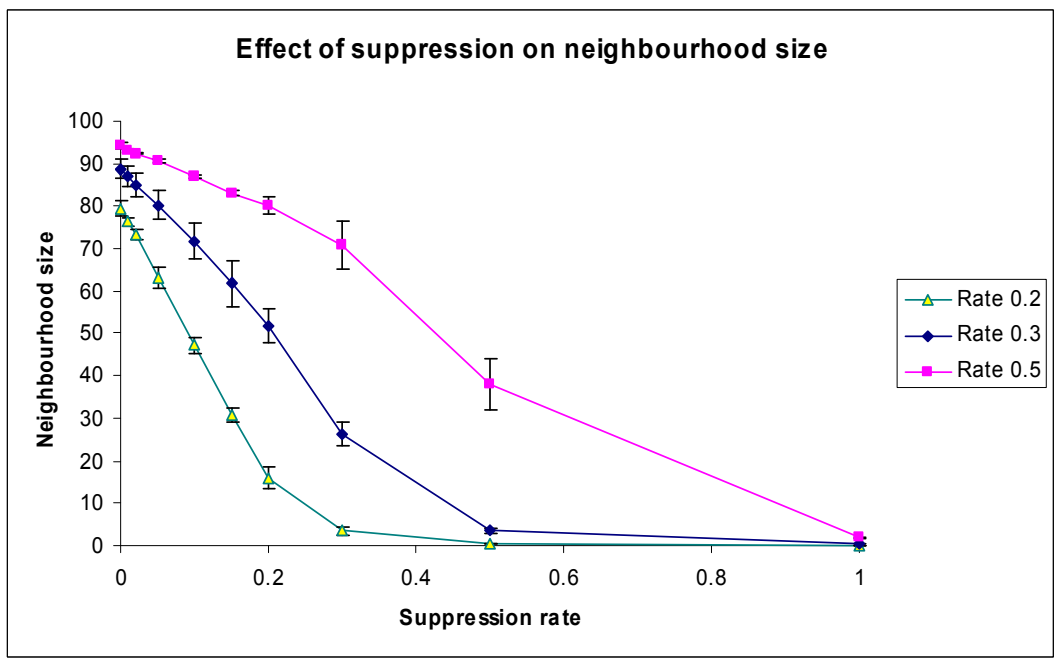

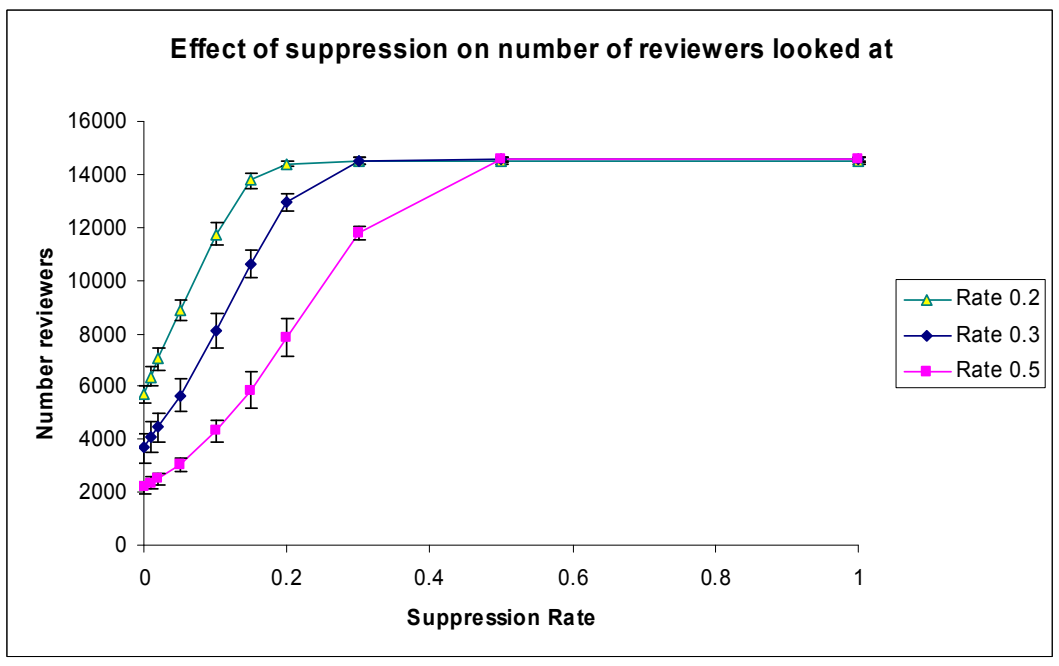

Figure 3: Effect of suppression rate on neighbourhood size and reviewers.

We then tested the effect of suppression on the AIS performance. Here we fixed the baseline rate at stimulation only (no suppression), and took measurements relative to this baseline. Again, it should be noted that the first graph shows prediction error (hence, a good result is low).

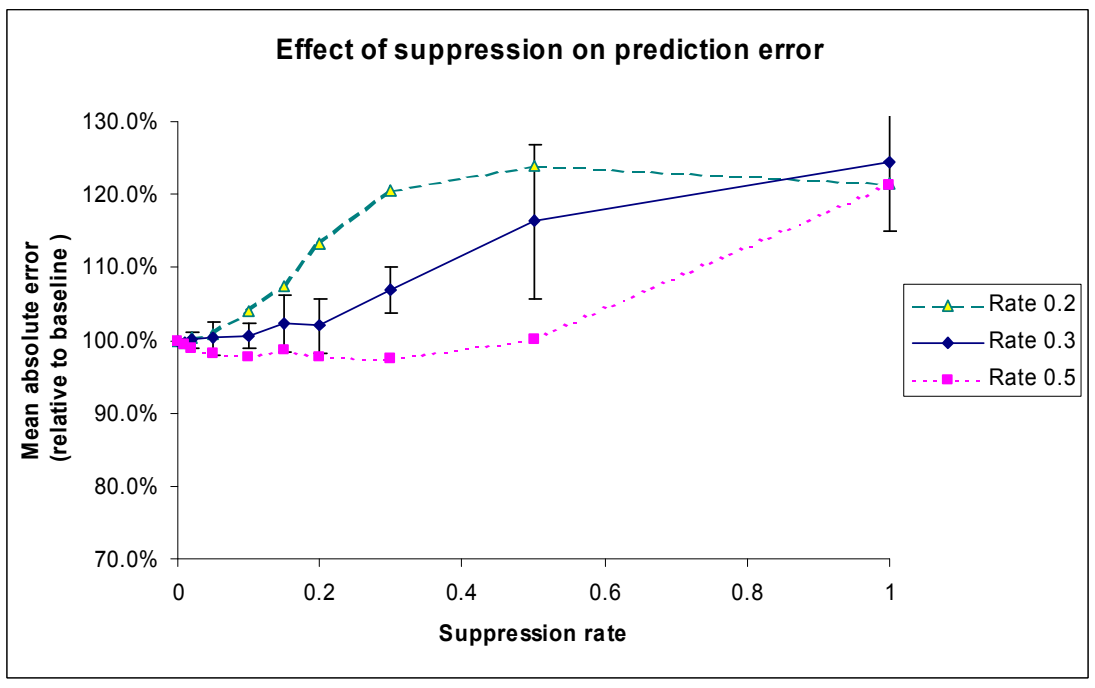

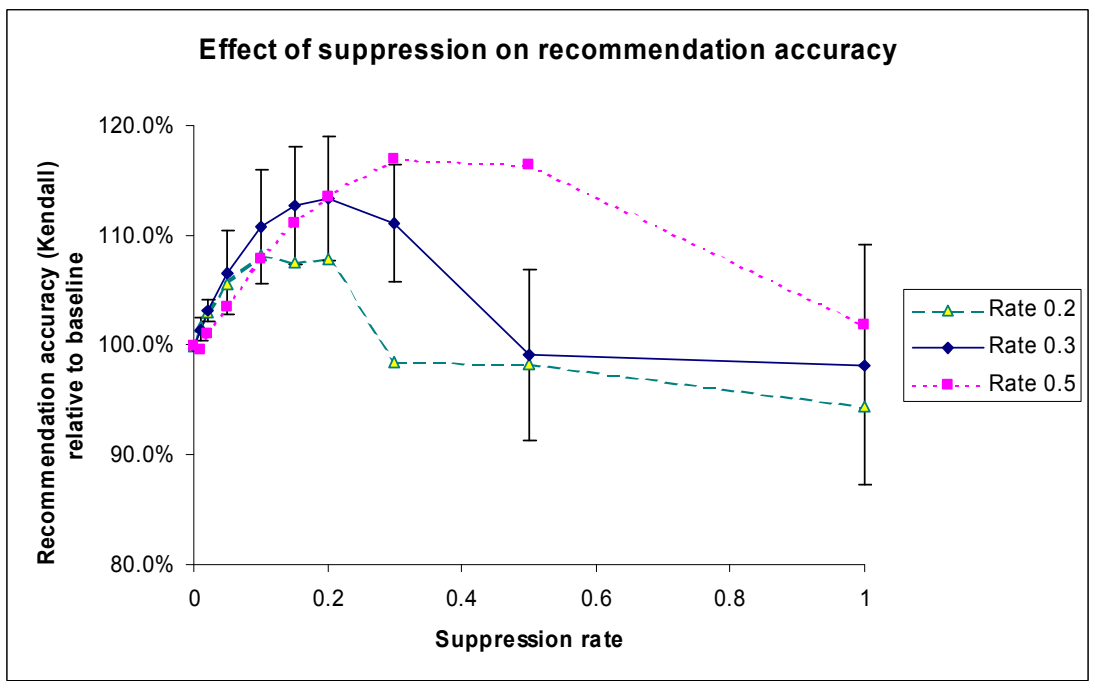

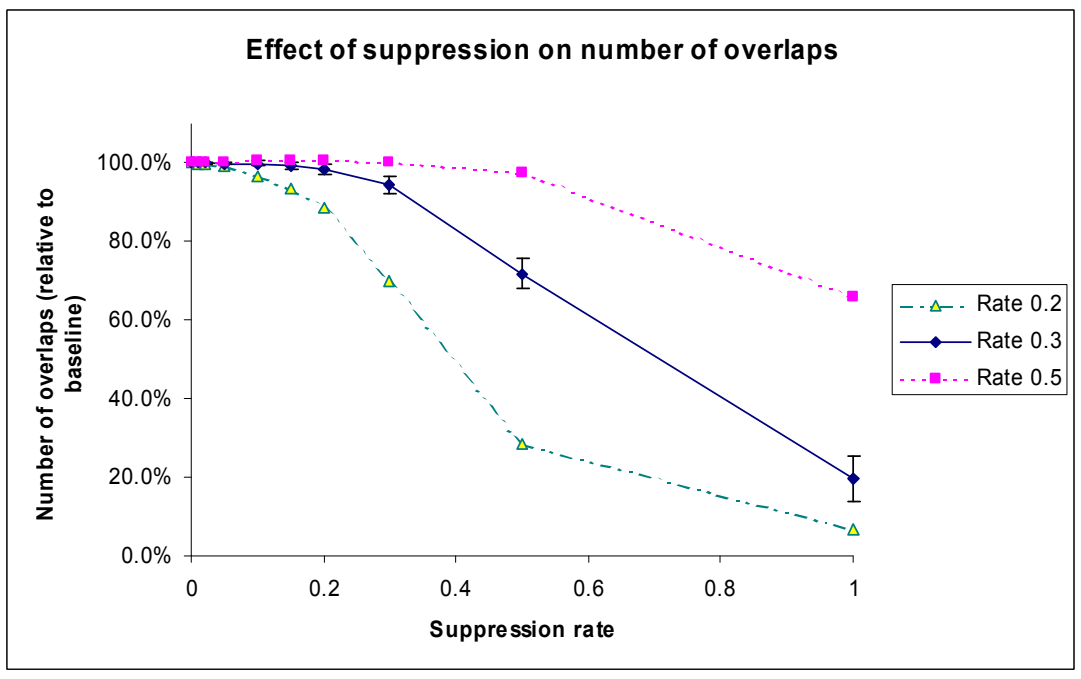

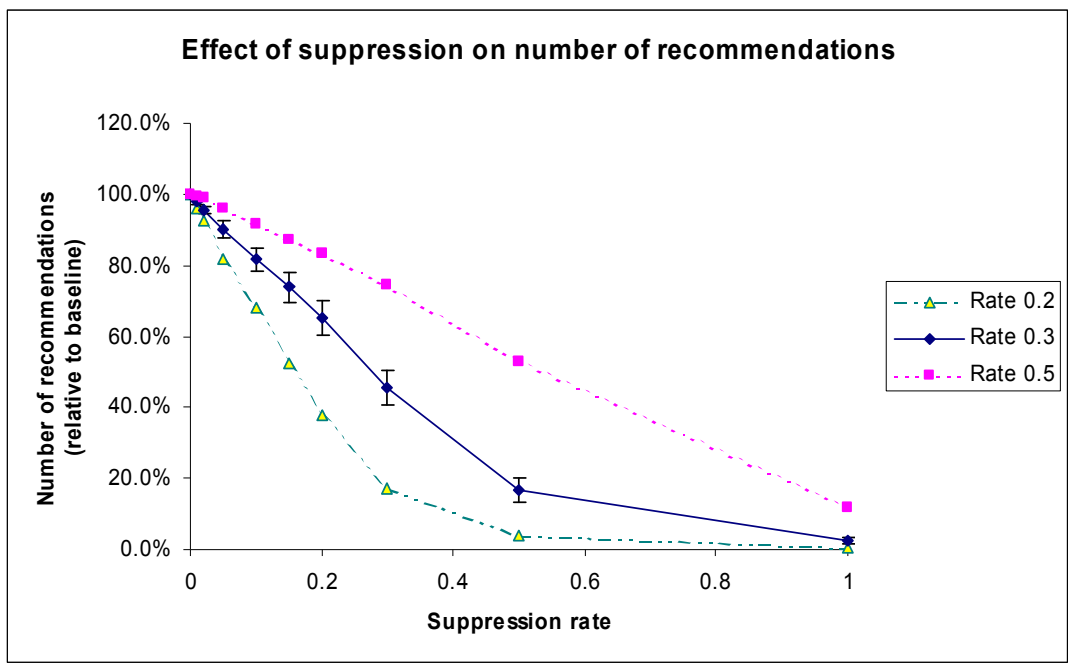

Figure 4: Effect of suppression rate on prediction and recommendation.

Again, the graphs show averaged results over five runs at each suppression rate. The bars show standard deviations (similar size bars for rates 0.2 and 0.5 have been omitted in the interests of clarity). At low levels of stimulation, prediction accuracy is not significantly affected. However recommendation accuracy is improved significantly (95% Wilcoxon). For instance, for 0.3 stimulation, rates from 0.05 to 0.2 gave a significantly improved performance. In actual terms, the Kendall measure rises from 0.5 to nearly 0.6. This means that the chance of any two randomly sampled pairs being correctly ranked has risen from 60% to 80%. Too much suppression had a detrimental effect on all measures.

## 4. CONCLUSIONS

It is not particularly surprising that the simple AIS performs similarly to the SP predictor. This is because they are, at their core, based around the same algorithm. The stimulation rate (in absence of any idiotypic effect) is effectively setting a threshold for correlation. This has both strengths and weaknesses. It has been shown that a threshold is useful in discarding the potentially misleading predictions of poorly correlated reviewers [10]. On the other hand, a rigid threshold means that one has to 'prejudge' the appropriate level to avoid both premature convergence and empty communities. Indeed, detailed examination of the individual runs showed that the AIS had a tendency to fill its neighbourhood either early or not at all. The setting of a threshold also means that sufficiently good antibodies are taken on a first come, first served basis. It is interesting to observe that such a strategy nevertheless seems (in these experiments) to provide a more constant level of overlaps, and better recommendation quality.

The richness of our AIS model comes when we allow interactions between antibodies. Early, qualitative experimentation with the idiotypic network showed antibody concentration rising and falling dynamically as the population varied. For instance, in the simple AIS, the concentration of an antibody will monotonically increase to saturation, or decrease to elimination, unaffected by the other antibodies. However, there is a delicate balance to be struck between stimulation and suppression. An imbalance may lead to a loss in population size or diversity. The graphs show that a small amount of suppression may indeed be beneficial to AIS performance, in particular recommendation. It is interesting to note that the increase in recommendation quality occurs with a relatively constant overlap size. At too high levels of suppression, it is harder to fill the neighbourhood, with consequent lack of diversity and hence recommendation accuracy.

We believe that these initial results show two things. Firstly, population effects can be beneficial for CF algorithms, particularly for recommendation; secondly, that CF is a promising new application area for artificial immune systems. In fact, we can widen the context, since the process of neighbourhood selection described in this paper can easily be generalized to the task of ad-hoc community formation.